\pdfoutput=1

\documentclass[11pt]{article}

\usepackage{acl}

\usepackage{times}
\usepackage{latexsym}
\usepackage{booktabs}
\usepackage{microtype}
\usepackage{amssymb}
\usepackage{pifont}
\usepackage{xcolor} 
\usepackage{graphicx}
\usepackage{placeins}
\usepackage{natbib}
\usepackage{tablefootnote}
\usepackage{longtable}
\usepackage{enumitem}
\usepackage{tikz}
\usepackage{booktabs}
\usepackage{pdfpages}

\usepackage[T1]{fontenc}

\usepackage{tabularx}
\usepackage{multirow}
\usepackage{xspace}
\usepackage{subfigure}
\usepackage{caption}
\usepackage{url}
\usepackage{amsmath}
\usepackage{amssymb}
\usepackage{amsfonts}
\usepackage{comment}
\usepackage{array}
\usepackage{multirow}
\usepackage{microtype}
\usepackage{pgfplots}
\usepackage{soul}
\usepackage{xcolor}
\usepackage{enumitem}
\usepackage{tikz}
\usepackage{cleveref}
\usepackage{pgf-pie}  

\usepackage[many]{tcolorbox}        
\usepackage{lipsum}

\tcbset{
    sharp corners,
    colback = white,
    before skip = 0.2cm,    
    after skip = 0.5cm      
}                           

\newtcolorbox{boxA}{
    boxrule = 0.5pt,
    fontupper=\small,
    colframe = black 
}

\usepackage[textsize=footnotesize]{todonotes}
\usepackage{linguex}
\usepackage{subcaption}
\alignSubExtrue

\usepackage[verbose]{newunicodechar}

\usepackage{inconsolata}

\usepackage{tcolorbox}

\tcbset{width=0.9\textwidth,boxrule=0pt,colback=red,arc=0pt,auto outer arc,left=0pt,right=0pt,boxsep=5pt}

\title{Beyond Good Intentions: When Does the Framing of Multilingual and Low-Resource NLP Research Become a Caricature?}

\author{Nedjma Ousidhoum$^\circ$, Noopur Zambare$^\diamond$, Mohamed Abdalla$^\diamond$\\
$^\circ$Cardiff University, $^\diamond$University of Alberta \\
Contact: \texttt{OusidhoumN@cardiff.ac.uk,mabdall2@ualberta.ca} 
}
 
\begin{document}
 \maketitle

\begin{abstract}

Building language technologies and conducting NLP research for low-resource languages---particularly when led by native speakers or involving participatory research practices---are often framed as means of addressing inequality, serving local communities, and, at times, contributing to \textit{decolonisation}. In this paper, we examine recently published NLP and ML papers, focusing on the narratives used to characterise multilinguality, low-resource languages, and underrepresented cultures. We propose a framework for analysing research framings and identify recurring rhetorical patterns that may hinder accountability and constrain equitable knowledge production for---and by---underserved communities.
We further assess the evidential basis of assertions regarding community benefit and find that such statements are often weakly supported or left unsubstantiated. Although community ownership and participation are frequently presented as key objectives, our analysis, supported by statistics from the ACL Anthology, suggests that research outputs more often prioritise resource creation and benchmarking---important but distinct goals---over evidence of broader structural change.
We conclude by offering practical recommendations to help authors, reviewers, and readers critically assess these assertions and avoid potentially misleading framings.\footnote{The data and guidelines used in this paper are available at \url{https://github.com/nedjmaou/Beyond_Good_Intentions}.}

\textcolor{red}{
\textbf{Warning:} This work highlights potentially problematic scholarly practices. To avoid misrepresentation or undue targeting, we use fictionalised, deliberately absurd, satirical examples for illustrative purposes. These examples are \textbf{not} intended as commentary on real publications or individuals, particularly students and early-career researchers.
}
\end{abstract}

\section{Introduction}

\begin{tcolorbox}[
    width=\linewidth,
    colback=red!5,
    colframe=black!60,
    boxrule=0.5pt
]
\centering
\textit{The English-speaking community has long been excluded from the technological systems that shape our digital sphere. Without immediate intervention, this imbalance risks undermining cultural continuity, democratic participation, economic opportunity, and the preservation of Europe's rich linguistic heritage. We invited a member of the affected community to contribute to our research. Through the creation of a benchmark, we take an important step toward linguistic justice, digital inclusion, and ensuring that English has a meaningful place in the future of artificial intelligence.}
\end{tcolorbox}

The preceding introduction is intentionally satirical. Yet replacing ``English'' with languages frequently discussed in multilingual and low-resource NLP makes the underlying rhetorical structure unexpectedly familiar. The satirical premise is not that underrepresented or endangered languages are undeserving of computational attention---they often are \cite{liu-etal-2022-always}. Rather, it is that multilingual NLP research can frame linguistic underrepresentation through narratives of urgency, moral responsibility, and anticipated social transformation that exceed what the technical contribution and available evidence can establish.

Clarifying research framings---what an algorithm, benchmark, dataset, or language technology is intended to do, why, by whom, and for whom---is essential for accountability \cite{schlichtkrull2023intended}. Such framings shape how technical work is justified, how impact is anticipated \cite{chamoun-etal-2025-social}, and how communities are represented \cite{liu-etal-2022-always}. In multilingual and low-resource NLP, these choices may be particularly consequential because technical artefacts are often positioned as interventions in larger linguistic, social, cultural, or political issues \cite{ousidhoum-etal-2025-building}. Papers may invoke goals such as language preservation, democratisation, decolonisation, community empowerment, inequality reduction, or digital inclusion \cite{bird-2020-decolonising}. At the same time, uneven language coverage, unequal access to language technologies, and historical marginalisation are legitimate concerns, and non-English NLP work may also face inequities in how it is developed and evaluated \cite{barkhordar2026nonenglishpapersreviewedfairly}, including negative review bias and expectations that exceed the scope of the claims being made. These concerns, however, do not in themselves establish a causal link between a technical contribution and a societal outcome; the mechanisms connecting technical work to broader social impacts therefore need to be clearly articulated.

\begin{figure}
    \centering
    \includegraphics[
        width=0.99\linewidth,
        height=0.25\textheight,
        trim=0 0 0 0,
        clip
    ]{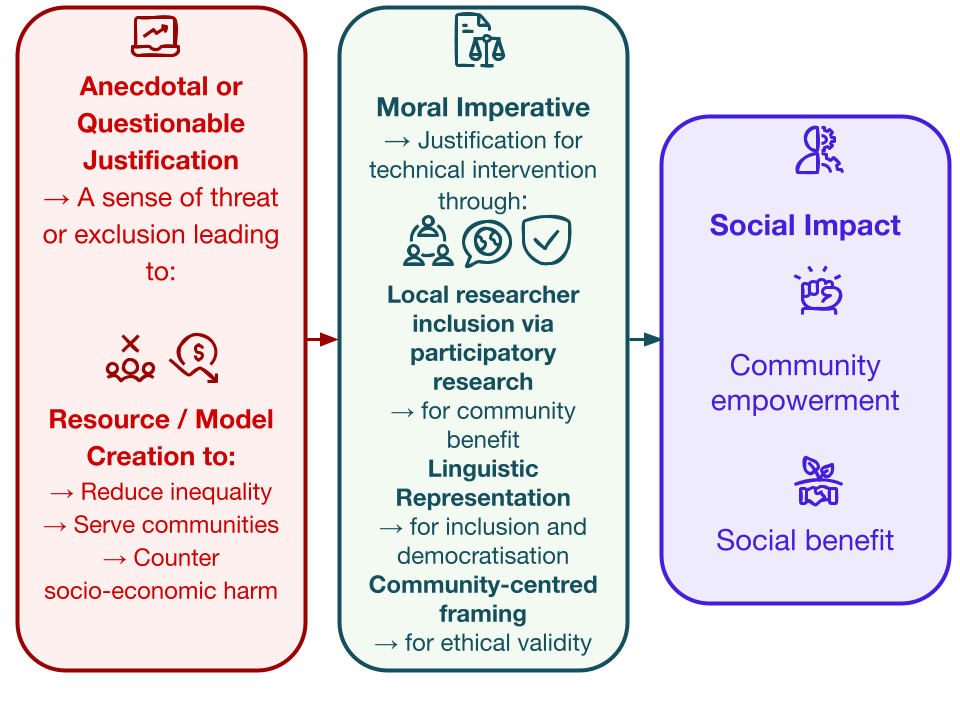}
    \caption{Commonly observed rhetorical and methodological patterns in research framings. Motivations (highlighted in red) are used to establish a moral imperative, the fulfillment of which (in green) is framed as leading to social impact (in blue).}
    \label{fig:causal_implications}
\end{figure}

In this paper, we investigate recurring framings in multilingual and low-resource NLP, focusing on how motivations, methodological choices, and anticipated societal outcomes are connected (see \autoref{fig:causal_implications}). Our focus is on whether research claims are proportionate to the evidence presented and whether the proposed mechanisms plausibly connect technical contributions to societal impacts. Specifically, we:
\begin{itemize}[noitemsep, nolistsep, leftmargin=*]
\item identify recurring narrative structures linking linguistic underrepresentation, technical intervention, and claimed societal benefits;
\item propose an analytical framework grounded in epistemic elements and research framings to examine how goals, methods, stakeholders, and anticipated outcomes are articulated;
\item apply this framework to multilingual and community-oriented low-resource NLP research, examining assumptions about participation, benefit, and impact;
\item investigate common patterns related to incentives, authorship, and the promotion of community-oriented multilingual NLP.
\end{itemize}

We find that social impact claims are often treated as natural consequences of resource creation, benchmark development, or model improvement. We also find that narratives of linguistic urgency sometimes appear to encourage assumptions about community needs, technological priorities, or anticipated benefits without sufficient corroborating evidence \cite{liu-etal-2022-always,adamu2023no,ferreira2024examining}. To encourage clearer alignment between technical contributions, supporting evidence, and societal claims, we conclude with an accountability checklist to help authors, reviewers, and readers critically evaluate multilingual and low-resource NLP papers and artefacts while maintaining proportionate expectations about their impact.

\section{Related Work}

Issues of accountability and responsible research have motivated several analyses of NLP practices, including studies of corporate influence in NLP and AI systems \cite{abdalla2021grey,abdalla-etal-2023-elephant,aitken2024collaboration}, as well as investigations of how research is framed and how such narratives may shape accountability. For instance, \citet{liu-etal-2023-responsible} reviewed summarisation papers and found that fewer than 15\% meaningfully engaged with responsible AI concerns. Similarly, \citet{schlichtkrull2023intended} formalised the notion of \textit{epistemic narratives} and analysed automated fact-checking papers, identifying frequent misalignments between goals and methods, underspecified or absent stakeholder definitions, and weak connections between these elements. \citet{chamoun-etal-2025-social} further extended this line of work by analysing more recent publications on automated fact-checking and hate speech, while also exploring automated approaches to research narrative analysis. Relatedly, \citet{subramonian2024understanding} examined how democratisation is conceptualised in NLP and ML research, finding limited engagement with established theories of democratisation despite frequent rhetorical invocation of the term.
Such studies reveal problematic research practices, broader trends in the field, and overlooked downstream risks \citep{birhane2022values}. They also provide guidance for future work by encouraging stronger alignment between goals and methods, while prompting reflection on concerns such as dual use \citep{leins-etal-2020-give,kaffee2023thorny} and overclaiming \citep{grodzinsky2012moral}. More broadly, ethical practices in AI, ML, and NLP have received considerable attention, including work on artefact documentation and transparency \cite{bender2011achieving,bender-tacl_a_00041,gebru2021datasheets,rogers-etal-2021-just-think,mohammad2022ethics}, as well as best-practice recommendations for responsible research \cite{hollenstein-etal-2020-towards}.

Despite growing attention to participation and equity in multilingual and low-resource NLP \cite{joshi-etal-2020-state,blasi-etal-2022-systematic,dogruoz-sitaram-2022-language}, work in this area has shed light on the general state of the field and common practices. This includes challenges in data collection \cite{yu-etal-2022-beyond}, the importance of involving speakers and communities whose languages are being studied \cite{bird-2020-decolonising,bird-2022-local,lent-etal-2022-creole,liu-etal-2022-always,mager2023ethical,bird2024centering}, corporate influence over language technologies and its implications for underrepresented linguistic communities \cite{held2023_nlp_coloniality_material,bird2025big}, and recent developments in large language model research \cite{mihalcea2024ai}.
However, relatively little attention has been paid to how research narratives and framings themselves shape methodological choices and justificatory claims in multilingual and low-resource NLP. Investigations of problematic epistemic narratives---such as an emphasis on numerical metrics and misalignment with social impact---have tended to focus on NLP and CL more broadly \cite{kogkalidis2025tables}. Meanwhile, critical discussions of building NLP artefacts for non-English, multilingual, and low-resource contexts have primarily focused on documentation and actionable recommendations \cite{bird2024centering,ousidhoum-etal-2025-building}. Efforts to make the review of non-English papers fairer have likewise addressed important aspects of the research process, including biased assumptions and a priori unrealistic expectations \cite{barkhordar2026nonenglishpapersreviewedfairly}. Nevertheless, we are not aware of any systematic investigation of problematic narratives, exaggerated claims, or research framings specifically in multilingual and low-resource NLP. Addressing this gap is the primary focus of this paper.

\section{Assessing Framings and Contributions}

\begin{table*}[!ht]
\centering
\small
\setlength{\tabcolsep}{4pt}
\begin{tabularx}{\textwidth}{XXX}
\toprule
\textbf{Ends (Goals)} &
\textbf{Means (Applications)} &
\textbf{Evidential Support} \\

\midrule

\begin{itemize}[leftmargin=*,noitemsep,topsep=0pt]
  \item \textbf{Serving communities:} developing artefacts for a specific community
    \item \textbf{Decolonisation:} addressing historical or structural inequalities
    \item \textbf{Reducing inequality:} promoting more equitable representation
    \item \textbf{Improving evaluation:} developing or enhancing benchmarks or metrics
    \item \textbf{Advancing knowledge production:} contributing new insights to research
    \item \textbf{Language revitalisation:} preserving, maintaining, or revitalising languages
    \item \textbf{Other:} Specified as needed 
\end{itemize}
&
\begin{itemize}[leftmargin=*,noitemsep,topsep=0pt]
\item \textbf{Improving LLM performance:} e.g., enhancing multilingual capabilities across tasks
\item \textbf{NLP applications:} e.g., translation, NER, transcription, and other applications
\item \textbf{LLM safety:} addressing safety and ethical considerations
\item \textbf{Improving LLM evaluation:} developing benchmarks/evaluation methods
\item \textbf{Supporting language learners:} e.g., developing tools for language tutoring
\item \textbf{Other:} specified as needed
\end{itemize}

&
\begin{itemize}[leftmargin=*,noitemsep,topsep=0pt]
    \item \textbf{Non-NLP scientific articles} evidence drawn from other academic domains
    \item \textbf{Newspaper articles or polls} media reports or collected public opinion
    \item \textbf{Prior NLP work:} evidence drawn from another NLP paper
    \item \textbf{Anecdotes} isolated examples or personal observations used as evidence
    \item \textbf{Expressions of threat:} claims of potential harm that are not substantiated by evidence
    \item \textbf{Opinions:} strong assertions lacking citation or empirical support
\end{itemize}
\\

\bottomrule
\end{tabularx}

\caption{Annotated epistemic elements and main categories used in our analysis. The \textit{ends} refer to the stated goals of the work, the \textit{means} refer to how authors intend to use their contributions to achieve those goals, and \textit{evidential support} refers to the evidence used to justify the claimed contribution.}
\label{tab:epistemic_elements}
\end{table*}

We select 100 papers addressing multilinguality broadly and low-resource languages in particular. We examine the claims made in these studies and how they contribute to broader research framings and narratives. Specifically, we analyse their abstracts and introductions, extracting relevant quotations and annotating them according to the elements defined below. Full annotation statistics are reported in the Appendix.

\subsection{Paper Selection}\label{subsec:paper_selection}
We first identified well-known and highly cited studies in the field, then expanded the pool through citation chaining, publications from prominent research consortia, and keyword-based searches of the ACL Anthology using terms such as \textit{multilingual}, \textit{low-resource}, \textit{reviv}, \textit{revival}, \textit{decolonization}, and \textit{decolon}. We manually screened the resulting pool to exclude studies that were not relevant to our research questions. We make the list of NLP papers publicly available on GitHub, while annotations are available upon request.

\subsection{Assessing Claimed Contributions}

Following \citet{schlichtkrull2023intended} and \citet{chamoun-etal-2025-social}, we examine passages pertaining to each paper's motivation (\textit{why}) and research focus (\textit{what}), primarily extracting quotations from the introductions and consulting abstracts when relevant information is missing. From these excerpts, we identify the \textit{means} (applications) and \textit{ends} (goals) of the work and characterise its \textit{narrative} (framing) based on the relationships among these elements. We additionally assess the extent to which stated goals are supported by evidence (i.e., \textit{evidential support} in \autoref{tab:epistemic_elements}).

\paragraph{Annotation Process} The annotation was conducted in two rounds by two expert annotators, both authors of this paper, following the recommendations of \citet{krippendorff2018content}. The process was iterative: we began with categories informed by an initial pilot study and introduced additional labels when needed. We employed a multi-label annotation scheme because individual passages could contain multiple relevant elements. The annotators achieved $>90\%$ agreement. Examples from the final label set are shown in \autoref{tab:epistemic_elements}.

\paragraph{Identifying the Means and Ends}
The \textbf{ends} refer to the purpose or ultimate aim of the work---that is, what the authors claim to accomplish. Based on our data, we identify the following main categories of ends:
\begin{itemize}[noitemsep,leftmargin=*]
\item \textbf{Serving communities} refers to goals explicitly directed towards benefiting \textit{a specific} group of people, for example by improving NLP technologies for a particular linguistic community.
\item \textbf{Decolonising} refers to goals framed in terms of contributing to the decolonisation of a region or marginalised group.
\item \textbf{Reducing inequality} refers, in contrast to serving communities, to \textit{broader goals} of improving representation, coverage, or equality beyond a specific community. For example, this includes generalising an observed gap from one language or language family to multiple groups.
\item \textbf{Advancing knowledge production} refers to contributing resources and insights that support scientific research.
\item \textbf{Improving evaluation} includes developing tools, benchmarks, or metrics that support the evaluation of NLP systems.
\item \textbf{Language revitalisation} refers to efforts to support the revitalisation of a language through data collection and NLP technologies.
\item \textbf{Other ends} include \textit{addressing environmental issues}, such as reducing the carbon footprint of NLP systems through greater efficiency; \textit{building powerful models}, including achieving state-of-the-art performance; and \textit{government monitoring}, where the stated aim involves monitoring government interests or activities in a particular region.
\end{itemize}

The \textbf{means}, or \textit{applications}, refer to how authors propose to use what is developed in a paper to accomplish a particular goal or end. Based on our data, we identify the following main categories of means:
\begin{itemize}[noitemsep,leftmargin=*]
\item \textbf{Improving LLM performance} refers to developing resources, methods, or models intended to improve the performance of LLMs for particular languages or tasks.
\item \textbf{NLP applications} include developing or improving technologies for specific tasks, such as machine translation, named entity recognition, transcription, and other applications.
\item \textbf{LLM safety and ethical considerations} refers to developing or applying methods intended to make language models safer or more appropriate for particular linguistic communities.
\item \textbf{Improving LLM evaluation} refers to developing methods for evaluating and comparing the capabilities of LLMs, particularly multilingual models, and identifying their limitations.
\item \textbf{Helping language learners} captures applications that use NLP technologies to support people learning one or more languages.
\item \textbf{Other} captures applications that do not fit within the categories above such as annotating or collecting data.
\end{itemize}

\paragraph{Narrative Annotation}
We identify epistemic narratives or framings based on the relationships among stated means, ends, and presented evidence (see \autoref{tab:epistemic_elements}). We define the following narrative categories:

\begin{itemize}[noitemsep,leftmargin=*]
\item \textbf{Knowledge Production}
A narrative in which the primary justification is scientific contribution. The stated goal is to advance understanding or support future research \textit{without} linking the work to broader societal outcomes.

\item \textbf{Explicit Pathway to Impact}
A narrative in which both means and ends are clearly articulated. The paper describes a concrete technical contribution (e.g., an automated language-processing task) and links it to a defined outcome (e.g., serving a specific community). This category excludes cases where the sole stated end is knowledge production.

\item \textbf{Vague Serving}
A narrative in which a technical contribution is presented as broadly serving a language community or population without a clearly articulated pathway to impact or supporting evidence.

\item \textbf{Saviourism / Decolonisation}
A narrative in which a resource or model is presented as saving an endangered language, reviving a historical language, or contributing to decolonisation.

\item \textbf{Tech Solutionism}
A narrative in which a technical resource or tool is presented as addressing inherently social, political, or structural problems (e.g., regional tensions) without supporting evidence or a demonstrated mechanism, potentially exaggerating the impact of the intervention.

\item \textbf{Tokenism}
A narrative in which limited acts of inclusion (e.g., involving annotators from underrepresented groups) are presented as meaningfully addressing broader structural inequalities in ways that may simplify or exaggerate the social impact of the work. This narrative is more common in promotional materials.
\end{itemize}

\section{The Risk of Exaggerated Claims in Multilingual and Low-Resource NLP}


We examine the evidential support and justificatory narratives used in the selected NLP papers. We find 42\% of the papers commonly rely on three forms of evidence: (i)\ perceived threats and anecdotal justifications, (ii)\ self-citation or weak evidential grounding, and (iii)\ broad claims regarding inequality reduction or community benefit that are weakly connected to the technical contribution.

\subsection{Perceived Threats and Anecdotal Justifications as Evidence}
We observe that a substantial number of claims rely on limited or insufficiently substantiated forms of justification. In several cases, assumptions, plausible narratives, or normative opinions are presented as scientific claims without clear empirical grounding. This distinction is important because plausible explanations can acquire the appearance of evidence even when the underlying empirical relationship has not been established \cite{brem2000explanation}.
More broadly, we observe recurrent narratives of urgency or threat used to justify technical work. These include claims that some language communities risk becoming ``second-class'' participants in technological progress, concerns about economic or societal ``exclusion from AI'', fears of linguistic disappearance, and arguments that language loss will lead to cultural homogenisation. While we acknowledge that some of these concerns are valid, the existence of a socially significant problem does not, by itself, demonstrate that a particular NLP system constitutes an effective response. Invoking such rationales without evidence specific to the population or context under discussion, or without establishing how the proposed intervention would mitigate the identified risk, can position technological researchers---even when they themselves belong to the broader community---as authoritative interpreters of community needs and interests \cite{adamu2023no}. 

For example, statements suggesting that the predominantly spoken nature of some languages makes them particularly suitable for technologies such as speech systems or multimodal large language models are sometimes made without empirical support, potentially representing local communities in \textit{agency-depriving ways} \cite{ferreira2024examining}. However, the relationship between a language's predominantly spoken nature and the effectiveness of speech-based technologies is not straightforward: in low-resource settings, limited training and evaluation data can constrain system performance. Therefore, the same scarcity of resources invoked to motivate such technologies may also limit their effectiveness, creating a tension between the proposed justification and the practical feasibility of the intervention. Moreover, the relationship between model capacity, data availability, and performance in low-resource languages is not straightforward, particularly in multilingual settings \cite{chang-etal-2024-multilinguality}.

\paragraph{The Limits of Citation-Based Justifications}
\begin{tcolorbox}[
    width=\linewidth,
    colback=red!5,
    colframe=black!60,
    boxrule=0.5pt
]
\textit{
Our claims are grounded in well-cited surveys that appear regardless of topical fit, alongside studies on language endangerment supported by a paper whose title was sufficiently alarming to obviate the need for further reading and a small number of self-citations arranged in mutually reinforcing sequence.}
\end{tcolorbox}
The presence of citations does not necessarily imply strong evidential support. In practice, feasibility support frequently consists of self-citation or references to a narrow set of commonly cited NLP papers (e.g., \citet{joshi-etal-2020-state}), rather than diverse or domain-specific evidence establishing a connection between the proposed technical intervention and the mitigation of stated risks, such as the economic marginalisation of Global Majority populations or linguistic homogenisation.
In some cases, references function primarily as rhetorical support for broad claims rather than as evidence demonstrating feasibility or a plausible causal mechanism. 
For example, concerns about language endangerment are often supported by references to prior NLP work, while leaving unspecified how the proposed technical contribution is expected to influence long-term language vitality \cite{bird-2020-decolonising}.

\subsection{The Gap Between Technical Contributions and Social Impact}
\begin{tcolorbox}[
    width=\linewidth,
    colback=red!5,
    colframe=black!60,
    boxrule=0.5pt
]
\textit{An information extraction system trained on English newswire data would simply report the F1 score. We, by contrast, report the F1 score \textbf{and} note that it represents a meaningful step toward the end of inequality in Europe.}
\end{tcolorbox}

Across the examined papers, `reducing inequality'' (48\%) and `serving communities'' (55\%) are frequently stated as end goals. To investigate how such claims are substantiated, we compare 10 papers addressing multilingual or low-resource languages with 10 others focusing exclusively on high-resource languages, primarily English and Chinese, across a range of representative NLP tasks, including NLU, NER, NLG, topic classification, multiple-choice QA, multitask language understanding, and language modelling. Specifically, we pair multilingual or low-resource papers with high-resource papers addressing the same task (e.g., a multilingual NER paper with an English or Chinese NER paper).
Despite comparable technical contributions, multilingual and low-resource papers in our sample more frequently invoked tech-solutionist narratives and broad societal outcomes---including disaster relief, addressing illiteracy, and reducing ethnic tensions---than their high-resource counterparts, including in cases where the primary contribution was a task-specific benchmark. By contrast, high-resource papers predominantly focused on technical objectives such as robustness evaluation and model performance. Notably, in the multilingual and low-resource papers, the pathway from technical contributions to claimed societal outcomes often remained underspecified. This gap was even more pronounced in institutional and project promotional materials, where we identified instances consistent with tokenism and saviourism across 17 collected online articles.
We emphasise that this asymmetry should be understood as a difference in scholarly rhetoric rather than as a difference in the value of research on any particular language. It may also reflect structural expectations and pressures placed on non-English NLP research during the review process, rather than researchers' choices alone \cite{barkhordar2026nonenglishpapersreviewedfairly}.

\subsection{Epistemic Risks of Exaggerated Claims}
\begin{tcolorbox}[
    width=\linewidth,
    colback=red!5,
    colframe=black!60,
    boxrule=0.5pt
]
\textit{Ultimately, Continental-EU-BERT is not merely a language model. It is a strike a decisive blow against centuries of computational inequity, colonial epistemology, and the lingering hegemony of English. 
It represents a meaningful step toward the empowerment of marginalised communities.
}
\end{tcolorbox}
We find that 11\% of the articles make unsupported claims about decolonisation or saviourism, while 14\% exhibit techno-solutionist narratives. Such rhetoric may obscure the limits of technical interventions by exaggerating what a resource or model can realistically achieve \cite{reiter2025we}, while potentially undermining the credibility of otherwise valuable technical contributions by weakening the connection between empirical findings and societal outcomes. It may also produce reductive representations of languages and their speakers, particularly when communities are framed primarily in terms of deficit, vulnerability, or technological absence \cite{bird-2022-local}. Even when the development of an artefact involves community members, an author's membership in a community does not, by itself, establish that they represent the experiences, priorities, or preferences of all its speakers. Differences in socioeconomic position, educational opportunities, geographic location, age, and access to technology can produce substantially different experiences within the same linguistic community \cite{warschauer2010new}. Statements made on behalf of a social group should therefore be distinguished from evidence concerning the perspectives of the particular groups intended to use or benefit from the proposed technology. Finally, such framing can create implicit moral imperatives around technical development, presenting language technologies as inherently necessary or ethically urgent despite limited evidence of their impact among underrepresented communities \cite{adamu2023no}.

\section{On the Risks of Constructing Implicit Moral Imperatives}
While we acknowledge the value of building technologies for underserved communities, framings that present technical intervention as inherently leading to greater inclusion of local researchers and increased linguistic representation, often reinforced through narratives of social benefit or empowerment \cite{mihalcea2024ai}, should be approached carefully. In our analysis, these framings and forms of evidential support (shown in green and red in \autoref{fig:causal_implications}, respectively) tend to legitimise technical intervention even when its necessity, feasibility, or expected impact remains questionable. In the following, we examine several prominent patterns through which such normative expectations are constructed.

\subsection{Local Researcher Inclusion and the Idealisation of Participatory Research}

Participatory research, particularly in the context of low-resource languages, has increasingly been framed as a pathway towards addressing longstanding inequities in NLP and ML research. While such approaches can foster more inclusive and community-centred practices, they are sometimes presented in idealised terms, with participation associated with broader outcomes such as the adequate representation of marginalised communities \cite{birhane2022power}. Such framings can implicitly treat participation as sufficient to address persistent power asymmetries, institutional hierarchies, and external pressures from dominant actors in academia and industry.

In some narratives, participatory research is portrayed as a collective responsibility through which researchers are expected to place their language or country ``on the map'', implicitly suggesting that greater representation can address deeper structural inequalities. However, approximately 20\% of the artefacts we analysed are unambiguously owned by Big Tech companies. This figure does not account for authors' institutional affiliations, although most papers include at least one author affiliated with a prestigious institution. These patterns suggest that community-oriented research can remain embedded within highly concentrated technological and institutional structures. Greater visibility, therefore, does not necessarily address persistent challenges such as underfunding, exploitative labour dynamics involving vendors and contractors in the Global South \cite{shahid2025think}, the potential misuse of language communities \cite{ousidhoum-etal-2025-building}, unequal access to computational resources, and broader structural barriers shaping research ecosystems \cite{png2022tensions}.

\paragraph{Framing Underrepresentation as a Threat}

Across multilingual and low-resource NLP papers, we observe recurring rhetorical patterns in the framing of motivations, methodological choices, and anticipated societal outcomes. These patterns often frame underrepresentation as a threat requiring technical intervention, as illustrated in \autoref{fig:causal_implications}. 
While we do not dispute the existence of disparities in available resources, we highlight cases in which rhetorical framing surpasses the empirical support and may encourage junior researchers who feel a sense of duty toward their language(s) to join projects without adequate compensation or recognition \cite{ousidhoum-etal-2025-building}. 

\paragraph{Collaboration vs.\ Sustained Engagement}

Although some papers emphasise collaboration with local communities or contributors, there is often limited evidence of sustained engagement beyond the initial study. In many cases, authors do not appear to continue publishing, raising questions about the continuity and depth of participatory claims.
Our analysis of papers in the ACL Anthology shows an increasing number of papers with at least 15 authors in recent years, rising from 9 in 2020 to 37 in 2021 and reaching 135 in 2025. Since 2015, authors of these papers account for approximately 2\% of all new authors, despite the papers themselves representing only 0.005\% of all publications (see Appendix).
In addition, among first-time authors of papers published since 2015, the probability of publishing again is close to zero. This raises questions about the extent to which participatory projects create sustained research engagement, particularly when they are presented as pathways for junior researchers from the Global South \cite{ousidhoum-etal-2025-building}.
Our intention is \textbf{not to discourage participation} in such projects, but to \textbf{encourage more careful communication} by established institutions and \textbf{more cautious interpretation} of participation opportunities by junior researchers from the Global Majority.

\subsection{Problem Framing and Pathways to Impact}

Papers motivated by linguistic revival (7\%) often invoke concerns about language extinction, digital exclusion, socioeconomic disadvantage, marginalisation, or safety risks. However, the relationships between these concerns and the proposed interventions often lack feasibility analyses, deployment evidence, or clearly articulated causal mechanisms grounded in relevant non-NLP research. Instead, such relationships are sometimes presented as self-evident, with limited consideration of the conditions under which technical interventions address the problems identified. Urgency is also reinforced through morally charged language, with aspirations such as democratisation ($n=6$) \cite{subramonian2024understanding}, community benefit, and inequality reduction ($>50\%$) used to connect technical contributions to broader societal outcomes without systematically establishing the pathways through which these outcomes would be achieved.

\paragraph{Representing Communities: Exceptionalism vs.\ Overgeneralisation}

An additional issue arises when a language or community is framed as uniquely underserved or exceptionally neglected without comparative baselines or supporting evidence, potentially contributing to its essentialisation. Stronger claims would ideally include comparisons across factors such as dataset coverage, benchmark availability, model performance, funding, or deployment constraints. Conversely, some work treats broad linguistic regions or large groups of languages as homogeneous, flattening meaningful linguistic, cultural, socioeconomic, and political differences \cite{adamu2023no,keleg-2025-llm}. More meaningful forms of empowerment require representations that are grounded in relevant social science research and empirical evidence and that account for variation within and across communities.

\paragraph{Underspecified Justifications for Artefact Creation}

Several papers justify new datasets, benchmarks, or models primarily by noting that the resource does not yet exist, while simultaneously linking its creation to broader societal impact. Although exploratory artefact creation is entirely legitimate, the motivation for creating a resource should not be conflated with demonstrated necessity: the absence of a benchmark or dataset does not, by itself, establish either scientific or practical need. Stronger motivations, where such claims are made, should include evidence of demand, utility, deployment relevance, or comparative benefits.

\paragraph{Insufficient Ethical Considerations}

Ethical concerns may also remain insufficiently addressed when underserved languages or community-centred approaches are invoked primarily as methodological framing, including relatively straightforward considerations such as annotation compensation \cite{ousidhoum-etal-2025-building}. This raises important questions regarding \textbf{whose priorities and values} are prioritised, which trade-offs are accepted, and how potential harms are identified and evaluated.

\section{Guidelines for Ethical Multilingual and Low-Resource NLP Research}

Based on our observations, broad claims about linguistic communities, technological needs, or societal outcomes can be problematic when supported by limited or narrowly scoped evidence. Recurring methodological limitations include:
\begin{itemize}[nolistsep,noitemsep]
\item limited sample sizes;
\item inadequate representativeness;
\item broad conclusions drawn from exploratory evidence;
\item insufficient acknowledgement of methodological limitations.
\end{itemize}

We therefore propose a checklist for stakeholders engaging with multilingual and low-resource NLP research, including papers, datasets, benchmarks, and industrial releases. The checklist focuses on recurring questions concerning research framing, evidentiary support, and implicit premises.
The questions below are intended to guide \textbf{authors}, \textbf{reviewers}, and \textbf{readers} when evaluating papers, blog posts, benchmarks, and product descriptions. They can help identify stereotypical or essentialist representations, as well as cases where claims about low-resource languages or underrepresented communities rely on assumptions that might be considered less acceptable in high-resource settings, such as those involving English, German, or French. They also draw attention to potentially asymmetric evidentiary expectations applied to non-English NLP research, which may contribute to gatekeeping. A useful guiding question across all roles is:

\begin{tcolorbox}[
width=\linewidth,
colback=blue!5,
colframe=black!60,
boxrule=0.5pt
]
\centering
\textbf{\textit{Would this claim be made in the same form in a high-resource language context? If not, what \textcolor{blue}{justifies} the difference?}}
\end{tcolorbox}

More broadly, the checklist aims to surface:
\begin{itemize}[noitemsep,nolistsep]
\item rhetorical overreach;
\item paternalistic framing;
\item techno-solutionist assumptions; and
\item unequal evidentiary standards.
\end{itemize}

\paragraph{Context and Language Ecology}
\begin{itemize}[noitemsep,nolistsep]
    \item Is exceptionalism being introduced without justification?
    \begin{itemize}[noitemsep,nolistsep]
        \item Would similar claims be made for a high-resource language?
        \item If not, are strong claims about a language or community necessary for motivating the work?
        \item Are common hype cycles or rhetorical patterns in the field being reproduced?    
        \end{itemize}

    \item Are low-resource language contexts treated as homogeneous, or is meaningful variation acknowledged?

    \item Do authors engage with relevant linguistic, sociolinguistic, anthropological, or regional scholarship where appropriate?

    \item Are local researchers meaningfully included in the work?
    \begin{itemize}[noitemsep,nolistsep]
        \item Has relevant non-English or regional scholarship been consulted?
    \end{itemize}

    \item Are technical systems implicitly framed as solving structural or societal problems?
\end{itemize}

\paragraph{Evidence and Data Quality}
\begin{itemize}[noitemsep,nolistsep]
    \item Are claims about language use, demographics, literacy, or infrastructure empirically grounded?
    \item Is language framed in essentialising terms (e.g., attributing outcomes to ``illiteracy'' or cultural deficit explanations)?
    \item Are dataset sources and provenance clearly documented?
    \item Has relevant domain expertise or community knowledge been consulted?
\end{itemize}

\paragraph{Stakeholder Engagement}
\begin{itemize}[noitemsep,nolistsep]
    \item Who are the intended users or affected communities?
    \begin{itemize}[noitemsep,nolistsep]
    \item How were their needs identified?
    \item Was there meaningful engagement, such as interviews, workshops, or field studies?
    \end{itemize}
    \item Are local priorities reflected in the research agenda?
    \item Is there evidence of engagement with relevant sociolinguistic or infrastructural contexts?
\end{itemize}

\paragraph{Inequality and Harm}
\begin{itemize}[noitemsep,nolistsep]
    \item Do research practices avoid extractive dynamics under the framing of ``inclusion'' or ``diversity''?
    \item Does the system shift dependency or centralise control?
    \item Are contributors (e.g., annotators) appropriately recognised and compensated?
    \item Is there space to challenge imported assumptions from dominant research hubs?
    \item Is there accountability beyond publication or initial deployment?
    \item Is evaluation conducted in relevant local contexts rather than only abstract or global settings?
    \item Are local practices and constraints adequately considered?
\end{itemize}

\section{Conclusion}
We examined how multilingual and low-resource NLP research is framed, with particular attention to equity and integrity. Our analysis suggests that, although these framings are often motivated by well-intentioned goals, the evidence supporting associated claims is frequently limited or unclear.
We highlight the need for stronger alignment between technical contributions and evidential support and provide practical guidance to help authors, reviewers, and readers critically assess such claims without imposing unjustifiably higher standards.

\noindent
We hope this work encourages reflection on how research practices, evaluation norms, and rhetorical conventions shape perceptions of research in the area.

\section*{Limitations}
This is not a comprehensive study but a starting point, as such it has a few limitations. The first limitation is that our analysed sample was of only 100 papers. While this sample size is within ranges of past work, it is not complete.

\noindent
Second, our analysis largely focused on abstracts and introduction. While we believe this decision is reasonable---as our initial more in-depth analyses of full papers did not yield more informative results---it is possible that some papers provided nuanced views or takes further down in the methodology or other sections.

\noindent
Lastly, we did not engage with the authors of papers to get their opinions or views. We explain our reasoning for this in the next section, but such interaction could have resulted in greater insights. 

\section*{Ethical Considerations}
The primary risk of this work is that it could inadvertently alienate, shame, or disproportionately affect particular authors, especially junior or underrepresented researchers. To mitigate this risk, our methodology is deliberately anonymised: we do not link findings to individual papers or provide direct quotes from the analysed works, instead using deliberately exaggerated and satirical examples.

\noindent
As stated in the introduction, while we identify potentially harmful narratives in some low-resource language research, we do not intend to suggest that low-resource languages are unworthy of research or technological development. Rather, our goal is to support the integrity of such efforts by advocating for more transparent, evidence-based, and practically grounded pathways to impact.

\paragraph{AI Use} We use privacy-preserving models to assist with proofreading, in line with the ACL guidelines.

\section*{Acknowledgments}
We thank anonymous reviewers for their helpful suggestions.
\newpage
\bibliography{anthology,custom}
\bibliographystyle{acl_natbib}
\newpage

\appendix
\section{Appendix: Statistics}

\begin{table}[ht]
\centering
\small
\begin{tabular}{l r}
\toprule
\textbf{Category} & \textbf{\%} \\
\midrule

\textcolor{blue}{\textbf{Ends}} & \\
Serve Communities & 55 \\
Decolonise & 2 \\
Fight Inequality & 48 \\
Language Revitalisation & 7 \\
Advancing knowledge production & 98 \\

\midrule

\textcolor{purple}{\textbf{Narratives}} & \\
Explicit Pathway to Impact & 85 \\
Knowledge Production & 99 \\
Saviourism/Decolonisation & 11 \\
Tech Solutionism & 14 \\
Vague Serving & 14 \\

\bottomrule
\end{tabular}
\caption{Distribution of main annotated ends and narrative categories across the analysed papers.}
\label{tab:annotation_counts}
\end{table}

\begin{table}[ht]
\centering
\begin{tabular}{lr}
\toprule
\textbf{Feasibility Support} & \textbf{Count} \\
\midrule
Previous NLP Paper & 76 \\
Opinion & 28 \\
Scientific Article/Poll & 9 \\
Sense of Threat & 11 \\
\bottomrule
\end{tabular}
\caption{Main types of feasibility support identified across the analysed papers.}
\label{tab:feasibility_support}
\end{table}

\begin{table*}[t]
\centering
\label{tab:acl_author_stats}
\small
\begin{tabular}{rrrrr}
\toprule
\textbf{Year} &
\multicolumn{1}{c}{\textbf{\% New Authors}} &
\multicolumn{1}{c}{\textbf{\(\leq 15\) Authors}} &
\multicolumn{1}{c}{\textbf{\(>15\) Authors}} &
\multicolumn{1}{c}{\textbf{\% Papers \(>15\)}} \\
\midrule
2000 & 0.1250    & 1253  &   1 & 0.0008 \\
2001 & 0         &  813  &   1 & 0.0012 \\
2002 & 0          & 1216  &   0 & 0      \\
2003 & 0          & 1186  &   0 & 0      \\
2004 & 0.0313    & 1922  &   2 & 0.0010 \\
2005 & 0         & 1260  &   1 & 0.0008 \\
2006 & 0.2941    & 2173  &   1 & 0.0005 \\
2007 & 0.0196    & 1598  &   3 & 0.0019 \\
2008 & 0         & 2231  &   1 & 0.0004 \\
2009 & 0         & 2232  &   1 & 0.0004 \\
2010 & 0.0250    & 3067  &   2 & 0.0007 \\
2011 & 0         & 2289  &   1 & 0.0004 \\
2012 & 0.0625    & 3390  &   1 & 0.0003 \\
2013 & 0.0455    & 2822  &   2 & 0.0007 \\
2014 & 0.0145    & 3634  &   5 & 0.0014 \\
2015 & 0         & 2955  &   3 & 0.0010 \\
2016 & 0.0107    & 4209  &  11 & 0.0026 \\
2017 & 0.0027    & 3481  &   6 & 0.0017 \\
2018 & 0.0343    & 4812  &  10 & 0.0021 \\
2019 & 0.0447    & 5065  &   9 & 0.0018 \\
2020 & 0.0129    & 7208  &  37 & 0.0051 \\
2021 & 0.0185    & 7107  &  41 & 0.0057 \\
2022 & 0.0185    & 8604  &  45 & 0.0052 \\
2023 & 0.0206    & 8980  &  52 & 0.0058 \\
2024 & 0.0264    & 12047 &  82 & 0.0068 \\
2025 & 0.0131    & 14751 & \textbf{135} & \textbf{0.0091} \\
\bottomrule
\end{tabular}
\caption{Evolution in the number of papers with fewer than and more than 15 authors in the ACL Anthology by year.}
\end{table*}

\end{document}